%% file: main.tex
\pdfoutput=1

\documentclass[11pt]{article}

\usepackage{acl}

\usepackage{times}
\usepackage{latexsym}

\usepackage[T1]{fontenc}

\usepackage[utf8]{inputenc}

\usepackage{microtype}
\usepackage{xspace}
\usepackage{subfigure}
\usepackage{booktabs}
\usepackage{tabularx}
\usepackage{longtable}
\usepackage{arydshln}
\usepackage{algorithm}
\usepackage{algorithmicx}
\usepackage{algpseudocode}
\usepackage{amsmath,amssymb,mathtools}
\usepackage{multirow}
\usepackage{bbm}

\newcommand{\cN}{\mathcal{N}}
\algnewcommand\algorithmicinput{\textbf{Input:}}
\algnewcommand\algorithmicoutput{\textbf{Output:}}
\algnewcommand\algorithmicparameter{\textbf{Parameters:}}
\algnewcommand\INPUT{\item[\algorithmicinput]}
\algnewcommand\OUTPUT{\item[\algorithmicoutput]}
\algnewcommand\PARAMETER{\item[\algorithmicparameter]}

\DeclareMathOperator{\Loss}{\mathcal{L}}
\DeclareMathOperator{\Dataset}{\mathcal{D}}
\newcommand{\method}{DCCL\xspace}

\title{Domain Confused Contrastive Learning for Unsupervised Domain Adaptation}

\author{Quanyu Long, Tianze Luo, Wenya Wang \and Sinno Jialin Pan \\
Nanyang Technological University, Singapore \\
\texttt{\{quanyu001, tianze001, wangwy, sinnopan\}@ntu.edu.sg}
}

\begin{document}
\maketitle

\begin{abstract}
\input{000abstract.tex}
\end{abstract}

\section{Introduction}
\label{sec:intro}
\input{001intro.tex}

\section{Preliminaries}
\label{sec:pre}
\input{002preliminaries}

\section{Method}
\label{sec:method}
\input{003method.tex}

\section{Experiments}
\label{sec:exp}

\input{004experiments.tex}

\section{Analysis}
\label{sec:ana}
\input{005analysis.tex}

\section{Related Work}
\label{sec:related}
\input{006related.tex}

\section{Conclusion}
\label{sec:conclusion}
\input{007conclusion.tex}

\section*{Acknowledgements}
\label{sec:ack}
\input{008acknowledgement}



\input{main.bbl}
\bibliographystyle{acl_natbib}

\clearpage
\appendix
\section{Datasets}
\label{sec:appendixA}
\input{010appendixA.tex}

\section{DANN}
\label{sec:appendixB}
\input{011appendixB.tex}



\end{document}

%% file: 000abstract.tex
In this work, we study Unsupervised Domain Adaptation (UDA) in a challenging self-supervised approach. One of the difficulties is how to learn task discrimination in the absence of target labels. Unlike previous literature which directly aligns cross-domain distributions or leverages reverse gradient, we propose Domain Confused Contrastive Learning (\method) to bridge the source and the target domains via domain puzzles, and retain discriminative representations after adaptation. Technically, \method searches for a most domain-challenging direction and exquisitely crafts domain confused augmentations as positive pairs, then it contrastively encourages the model to pull representations towards the other domain, thus learning more stable and effective domain invariances. We also investigate whether contrastive learning necessarily helps with UDA when performing other data augmentations. Extensive experiments demonstrate that \method significantly outperforms baselines.

%% file: 001intro.tex
Pre-trained language models \citep{Devlin2019BERTPO,Liu2019RoBERTaAR,Yang2019XLNetGA} have yielded considerable improvements
with datasets drawing from various sources. 
However, the lack of portability of language model to adapt to a new textual domain remains a central issue \citep{gururangan2020don}, especially when the training set and testing set do not follow the same underlying distribution (changing of topic and genres) - usually referred to as domain shift.
In this paper, we focus on studying Unsupervised Domain Adaptation (UDA). UDA aims at designing adaptation algorithms that attempt to generalize well on the target domain by learning from both labeled samples from the source domain and unlabeled samples from the target domain. Studying UDA fits real-world scenarios since labeled data in the target domain is usually absent. 
Moreover, advances in UDA will also help out-of-distribution generalizations \citep{ramponi2020neural,krueger2021out}.

\begin{figure}[t]
    \centering
    \includegraphics[width=0.99\columnwidth]{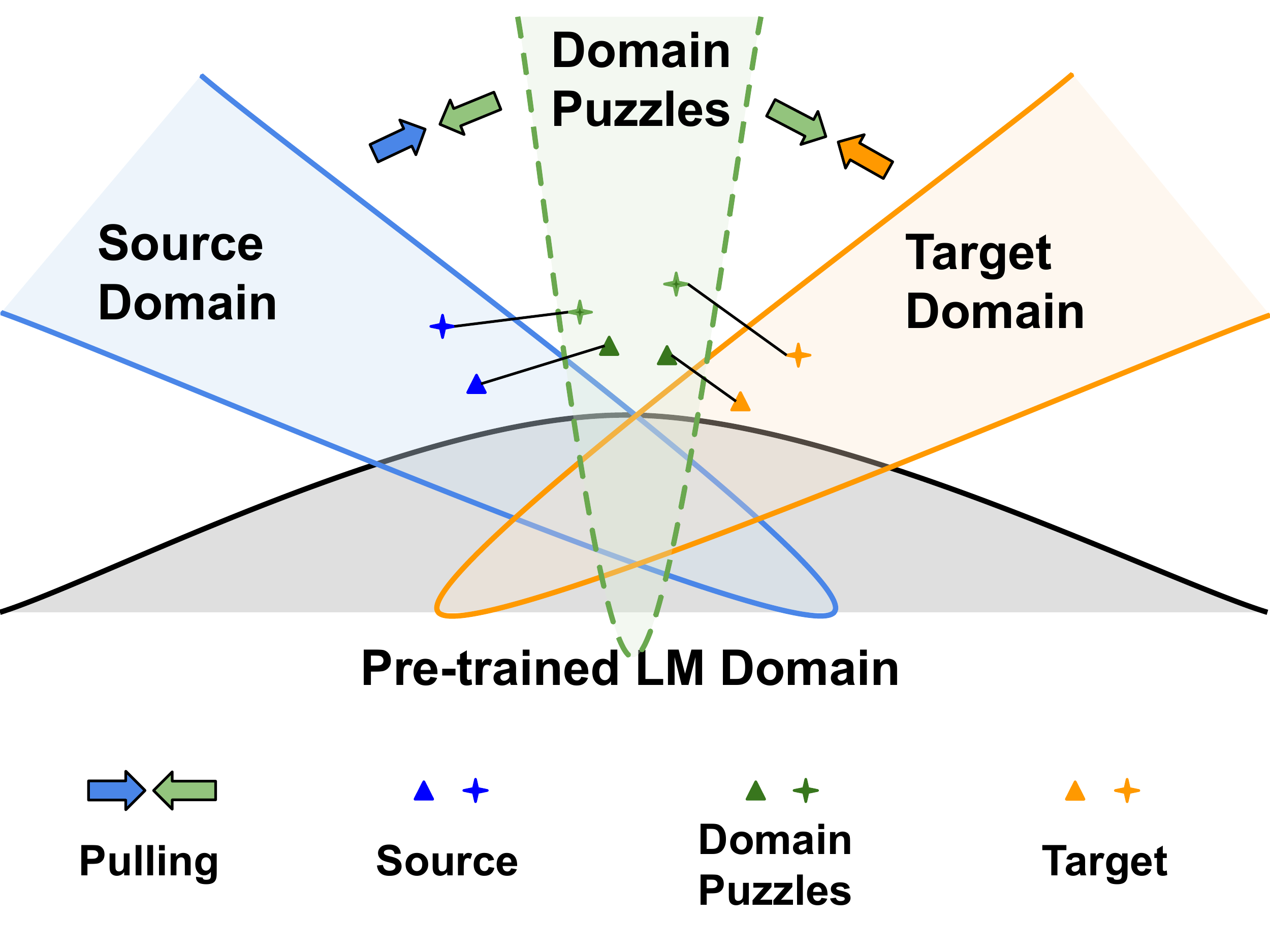}
    \caption{Domain puzzles that are domain-confused and overlook domain-related information, could be regarded as lying in an intermediate domain that aims to pull source and target samples closer to each other and bridge the two domains by learning domain invariant representations.}
    \label{fig:moti}
\end{figure}

Extensive algorithms have been proposed to mitigate the domain shift problem, for example, domain adversarial neural network (DANN) \citep{ganin2016domain} and distribution matching \citep{zhuang2015supervised}.
For DANN, the training process of joint optimization is unstable, requiring extensive effort to tune the hyperparameters \citep{shah2018adversarial,du2020adversarial,karouzos2021udalm}. 
As for distribution matching, it is very difficult to preserve the discriminative power of the model on the target task while trying to perform instance level alignment \citep{saito2017asymmetric,lee2019drop}. 
To this end, it is essential to develop stable and effective solutions to learn domain invariance and instance-wise matching for UDA.

Recent advances in self-supervised learning (SSL), such as contrastive learning (CL), have been proven effective at instance level by leveraging raw data to define surrogate tasks that help learn representations \citep{chen2020simple,khosla2020supervised,he2020momentum,chen2020improved}. 
CL benefits from treating instances as classes and conducting data augmentations to generate positive instance pairs.
Regarding CL for UDA, previous works mark cross-domain images with the same labels (for example, real and cartoon dogs) as the positive pairs in contrastive loss \citep{wang2021cross,park2020joint}. However, such methods are not applicable to NLP tasks because of the massive semantic and syntactic shifts between two cross-domain sentences.
Besides, from the domain adaptation perspective, constructing cross-domain positive samples and aligning domain-agnostic pairs have received less emphasis in related literature, since previous works focus on designing label preserving text transformations, such as back-translation, synonym, dropout and their combinations \citep{qu2021coda,gao2021simcse}.

Confronting with limitations mentioned above, we propose the concept of domain puzzles which discard domain-related information to confuse the model, making it difficult to differentiate which domain these puzzles belong to.
Instead of directly seeking matched sentences across the source and target domains which is infeasible, we propose to pull the source (target) data and its corresponding domain puzzles closer to reduce the domain discrepancy, as shown in Fig.~\ref{fig:moti}. 
A simple idea to craft domain puzzles is to mask domain-specific tokens. However, token-level operations are too discrete and non-flexible to reflect the complex semantic change of natural languages. Hence, we aim to seek better domain puzzles that retain high-confidence predictions and task-discriminative power in the representation space for each training instance.
In this paper, we propose Domain Confused Contrastive Learning (\method) to encourage the model to learn similar representations for the original sentence and its curated domain-confused version with contrastive loss.
More specifically, we synthesize these domain puzzles by utilizing adversarial examples \citep{zhu2020freelb,jiang2020smart}. The algorithm will search for an extreme direction that roughly points to the opposite domain and produces most domain-challenging puzzles.
We encourage the model to encode original and domain-confused samples closer, gradually pulling examples to the domain decision boundary as training progresses via CL, thus learning the domain invariance.
Furthermore, in order to investigate whether CL necessarily benefits UDA,
we conduct experiments and find that constructing domain puzzles as paired positive samples is favorable for UDA, however, other data augmentation methods such as back translation \citep{sennrich2016improving,edunov2018understanding} do not have the same effect. The experiment results show that the proposed \method significantly outperforms all the baselines. We also conduct quantitative experiments to measure the domain discrepancy after adaptation demonstrating that \method can decrease the divergence between domains in a self-supervised way.
Overall, the paper makes the following contributions:
\begin{itemize}
    \item First, a new concept of domain puzzles is put forward. We propose to craft the domain puzzles via domain-confused adversarial attack;
    \item Second, we propose \method, which is able to pull source and target samples closer to the crafted domain puzzles. The \method is capable of reducing domain shift and learning domain invariance;
    \item Third, experiments demonstrate that the proposed \method surpasses baselines with a large margin. We also conduct analyzing experiments to verify the effectiveness.
\end{itemize}

%% file: 002preliminaries.tex
\begin{figure*}[t]
\centering
\includegraphics[width=0.99\textwidth]{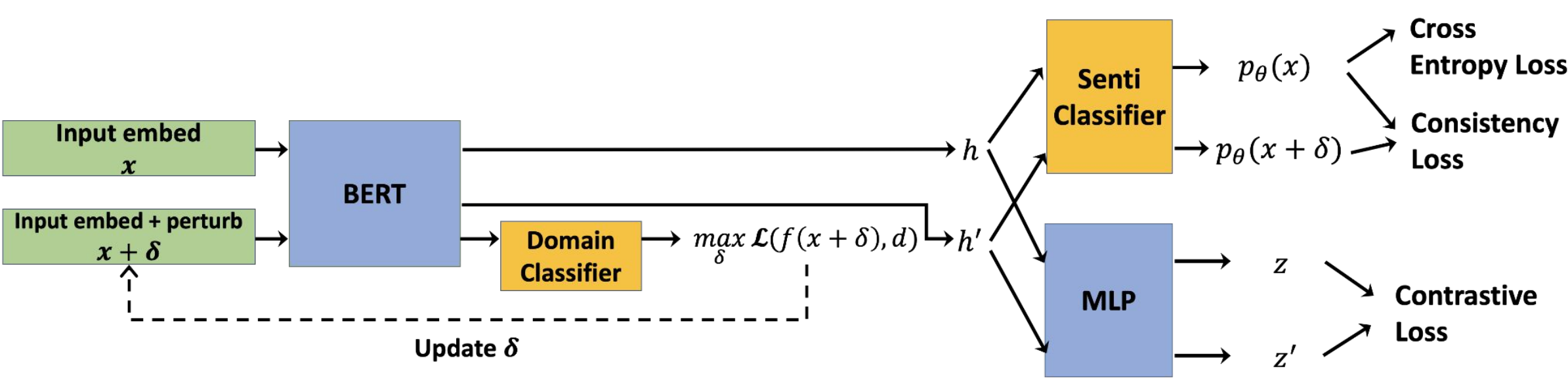}
\caption{Overview of the framework of the proposed method \method.}
\label{fig:model}
\end{figure*}

\subsection{Unsupervised Domain Adaptation}
\textbf{Problem Setup} Suppose we have access to a source dataset with $n$ labeled data points $\Dataset^{S}=\{x_{i},y_{i}\}_{1,...,n}$ sampled i.i.d. from the source domain, and a target dataset with $m$ unlabeled points $\Dataset^{T}=\{x_{j}\}_{1,...,m}$ sampled i.i.d. from the target domain, where $x_{i},x_{j}$ are sequences of tokens, $y_{i}$ is the class label for $x_i$. For in-domain training with labeled training instances, the model aims to learn a function $f(x;\theta_{f},\theta_{y}):x\rightarrow C$. $\theta_{f}$ is the parameter of the deep neural network encoder (e.g., pretrained language model), $\theta_{y}$ denotes parameters that compute the network’s class label predictions, and $C$ is the label set. The model is learned with the following objective:
\begin{equation}
    \min_{\theta_{f},\theta_{y}} \sum\nolimits_{(x,y)\sim \Dataset^{S}}[\Loss(f(x;\theta_{f},\theta_{y}),y)].
\end{equation}

However, for Unsupervised Domain Adaptation (UDA), the goal of the adaptation algorithm is to learn a discriminative classifier from the source domain, which at the same time could generalize well on the target domain by leveraging unlabeled target data and learning a mapping between source and target domains. It is generally acknowledged that the discrepancy between two datasets (domain shift) can be reduced by aligning two distributions \citep{ben2007analysis,ben2010theory}. The methods that learn domain invariant features for domain alignment include KL divergence \citep{zhuang2015supervised}, Maximum Mean Discrepancy (MMD) \citep{gretton2012kernel}, and Domain Adversarial Neural Network (DANN) \citep{ganin2016domain} (details of DANN can be found in Appendix. \ref{sec:appendixB}). DANN suffers from a vanishing gradient problem \cite{shen2018wasserstein}, and the training process is unstable \citep{shah2018adversarial,du2020adversarial}. Hence, more efficient and stable algorithms are essential for UDA \citep{wu2019domain}.

\subsection{Adversarial Training}
Adversarial training with perturbations has been shown to significantly improve the performance of the state-of-the-art language models for many natural language understanding tasks \citep{madry2018towards,zhu2020freelb,jiang2020smart,pereira2021targeted}. The algorithm generally considers adversarial attacks with perturbations to word embeddings and minimizes the resultant adversarial loss around the input samples.
In a single domain, adversarial training \citep{goodfellow2014explaining} is an inner max, outer min adversarial problem with the objective:
\begin{equation}
    \min_{\theta_{f},\theta_{y}} \sum\nolimits_{(x,y)\sim \Dataset}[\max_{\delta}\Loss(f(x+\delta;\theta_{f},\theta_{y}),y)].
    \label{adv}
\end{equation}
With (\ref{adv}), the standard adversarial training can also be regularized using virtual adversarial training \citep{miyato2018virtual}, which encourages smoothness in the embedding space. 
The $\alpha_{adv}$ controls the trade-off between the two losses, usually set to be $1$.
\begin{equation}
\begin{split}
    \min_{\theta_{f},\theta_{y}}&\sum\nolimits_{(x,y)\sim \Dataset}\big[\Loss(f(x;\theta_{f},\theta_{y}),y)+\alpha_{adv} \\
    &\max_{\delta}\Loss(f(x+\delta;\theta_{f},\theta_{y}),f(x;\theta_{f},\theta_{y}))\big].
\end{split}
\label{vat}
\end{equation}

For (\ref{adv})(\ref{vat}), the inner maximization can be solved by Projected Gradient Decent (PGD) \citep{madry2018towards} with an additional assumption that the loss function is locally linear. A following iteration can approximate the adversarial perturbation $\delta$:
\begin{align}
    &\delta_{t+1}=\Pi_{\lVert \delta \rVert _{F}\leq \epsilon}(\delta_{t}+\eta\frac{g^{adv}_{y}(\delta_{t})}{\lVert g^{adv}_{y}(\delta_{t}) \rVert}_{F}), \\
    &g^{adv}_{y}(\delta_{t})=\nabla_{\delta}\Loss(f(x+\delta_{t};\theta_{f},\theta_{y}),y),
\end{align}
where $\Pi_{\lVert \delta \rVert _{F}\leq \epsilon}$ performs a projection onto the $\epsilon$-ball. The advantages of PGD lie in that it only relies on the model itself to produce diverse adversarial samples, enabling the model to 
generalize better to unseen data.

%% file: 003method.tex
In this section, we focus our discussions on the proposed Domain Confused Contrastive Learning (\method) under a sentiment classification scenario. The overall framework of our method is illustrated in Fig.~\ref{fig:model}. The model will take source labeled and target unlabeled sentences as input. It will then augment the input data with domain puzzles by fabricating adversarial perturbations. With the augmented data, the next step produces a hidden representation for each instance with an encoder which will be further used to produce three losses to train the entire model, namely sentiment classification loss, contrastive loss and consistency loss.

\subsection{Crafting domain puzzles}

\begin{figure}[t]
    \centering
    \includegraphics[width=0.95\columnwidth]{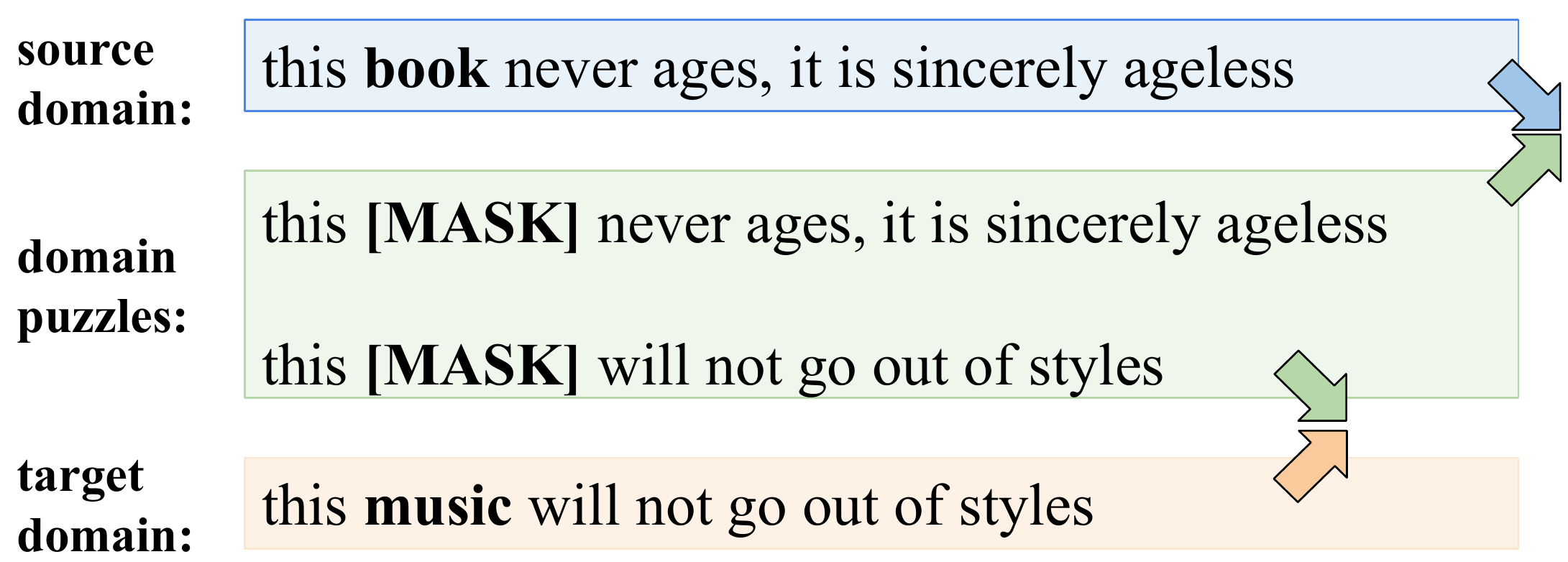}
    \caption{Two sentences sampled from Book and Music reviews. Alternatively, we can match original sentences with its degraded masked versions.}
    \label{fig:mask}
\end{figure}

For UDA, \citet{saito2017asymmetric} mentions that simply matching the distributions cannot ensure high accuracy on the target domain without the target labels. Moreover, it may cause negative transfer, deteriorating knowledge transfer from source domain to the target domain
\citep{wang2019characterizing}. Even if the matched sentences have the same label, due to huge syntactic and semantic shift, instance-based matching strategies that align examples from different domains will introduce noises for pre-trained language models, for example, aligning source domain and target domain sentences in Fig.~\ref{fig:mask}.

Alternatively, we can locate and mask domain-specific tokens which are related to sentence topics and genres. Since sentences in the green box of Fig.~\ref{fig:mask} become domain-agnostic, we refer to those domain-confused sentences (one cannot tell which domain these sentences belong to) as domain puzzles. Matching distributions between the source domain and the domain puzzles, as well as the target domain and the domain puzzles, will also make language models produce domain invariant representations.

However, the domain-specific tokens are not always evident, due to the discrete nature of natural languages, it is challenging to decide correct tokens to mask without hurting the semantics especially when the sentences are complicated\footnote{Masking is not our focus in this paper. More detailed implementation can be found in section \ref{sec:baselines}. We will investigate how to extract better domain tokens in our future work.}.
Hence, we seek domain puzzles in the representation space and introduce adversarial perturbations, because we can rely on the model itself to produce diverse but targeted domain puzzles. Note that the purpose of adversarial attack here is not to 
enhance the robustness, but to construct exquisitely produced perturbations for a better domain invariance in the representation space.

To generate domain-confused augmentations, we adopt adversarial attack with perturbations for domain classification. The loss for learning a domain classifier with adversarial attack can be specified as follows:
\begin{align}
    \Loss&_{{\rm domain}}=\Loss(f(x;\theta_{f},\theta_{d}),d)+ \notag \\
        &\alpha_{adv}\Loss(f(x+\delta;\theta_{f},\theta_{d}),f(x;\theta_{f},\theta_{d})), \\
    \delta&=\Pi_{\lVert \delta \rVert _{F}\leq \epsilon}(\delta_{0}+\eta\frac{g^{adv}_{d}(\delta_{0})}{\lVert g^{adv}_{d}(\delta_{0}) \rVert}_{F}), \label{delta}
\end{align}
where $\delta_{0}$ is the initialized noise, $\theta_{d}$ is the parameter corresponding to the computation of the domain classification, and $d$ is the domain label. Due to additional overhead incurred during fine-tuning large pre-trained language models, the number of iterations for perturbation estimation is usually 1 \citep{jiang2020smart,pereira2021targeted}, as shown in Eq. \ref{delta}. We synthesize the perturbation $\delta$ by searching for an extreme direction that perplexes the domain classifier most in the embedding space, and $f(x+\delta;\theta_{f})$ is the crafted domain puzzles encoded by the language model.

\begin{figure}[t]
\centering
\subfigure[Postive sampling]
{
\centering
\includegraphics[width=0.9\columnwidth]{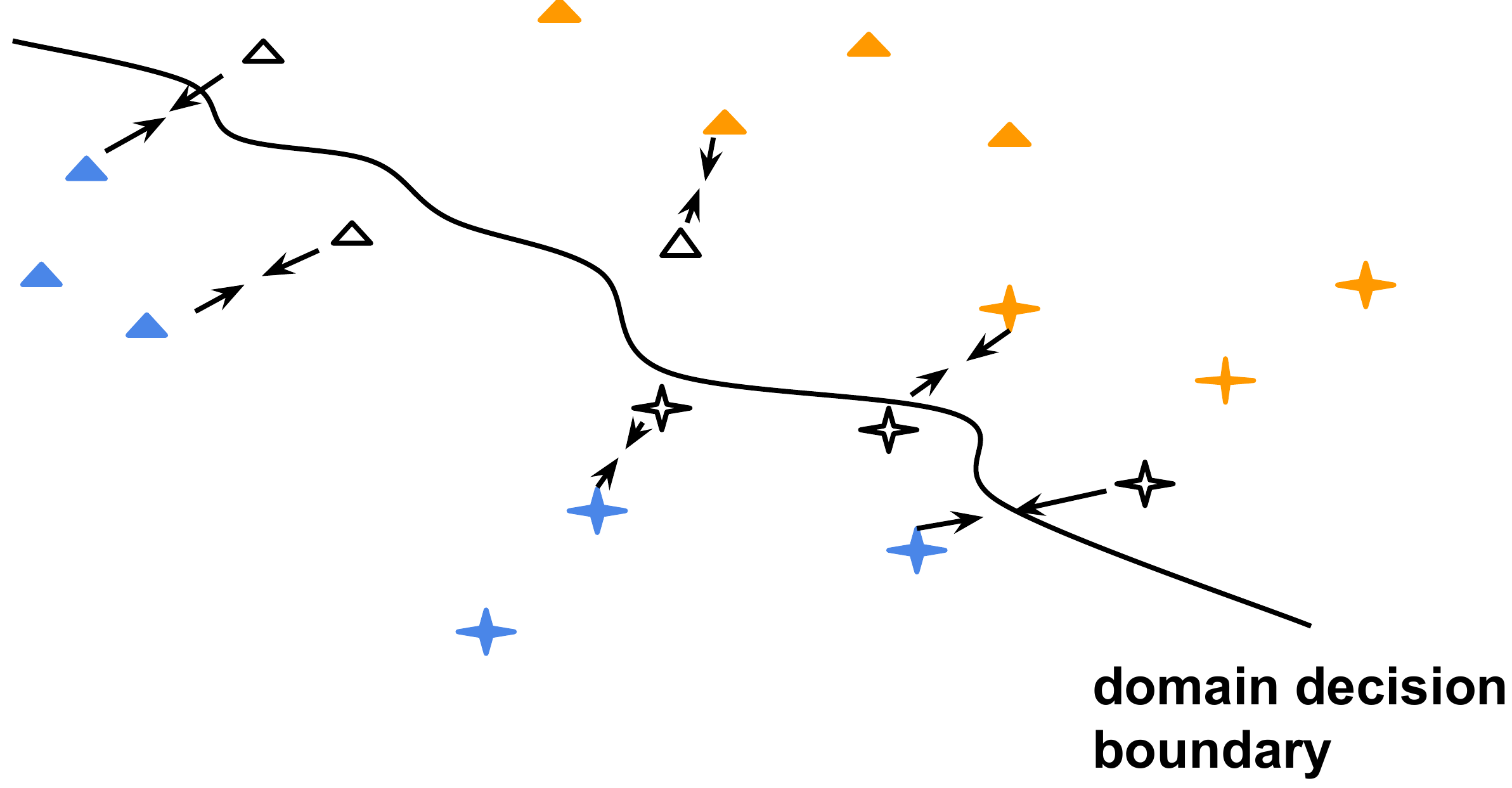}
}
\subfigure[Negative sampling]
{
\centering
\includegraphics[width=0.95\columnwidth]{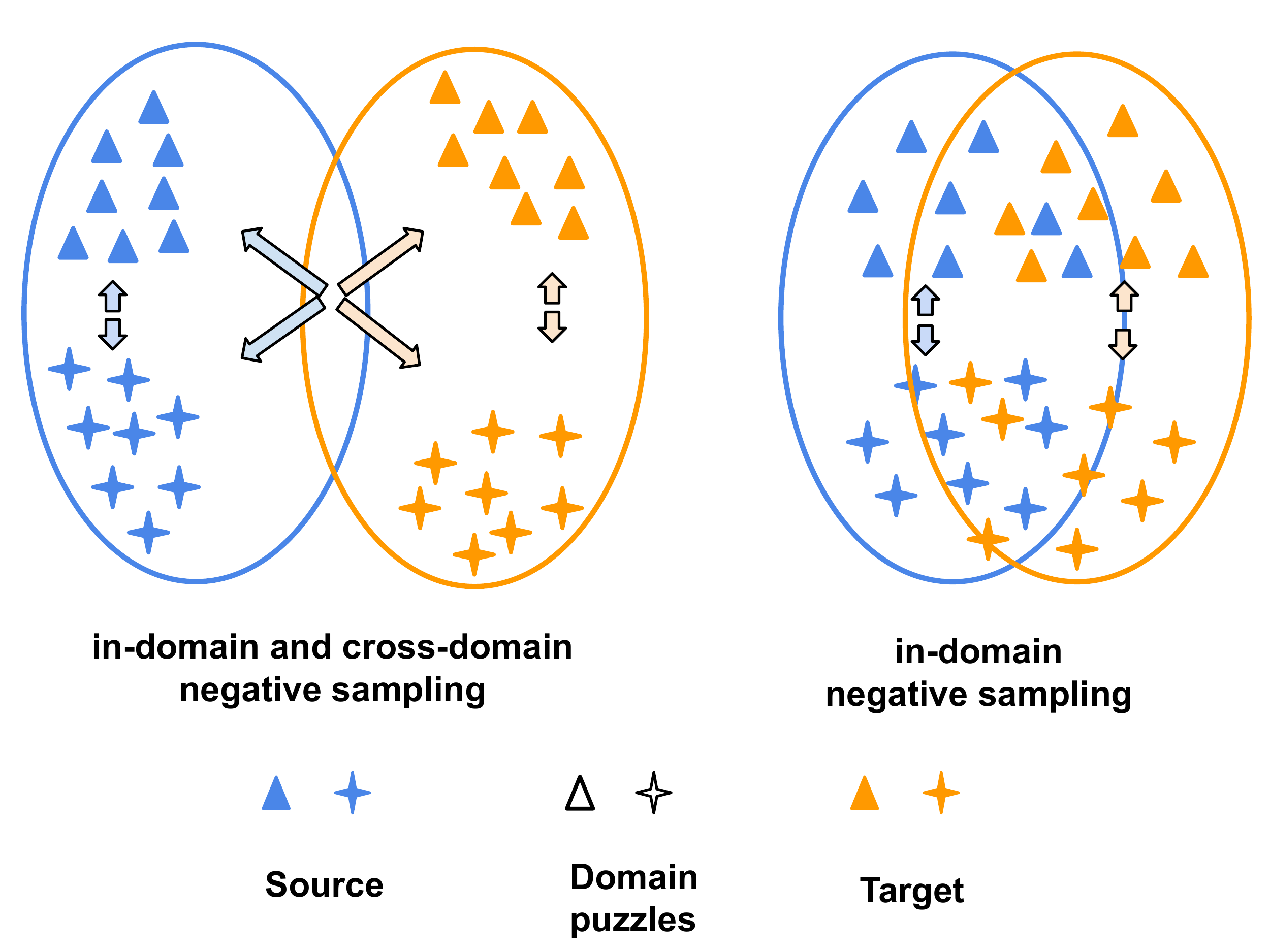}
}
\caption{Proposed adaptation sampling method}
\label{fig:sampling}
\end{figure}

\subsection{Learning invariance with domain puzzles}
After acquiring domain puzzles, simply applying distribution matching will sacrifice discriminative knowledge learned from the source domain \citep{saito2017asymmetric,lee2019drop}, and instance-based matching will also overlook global intra-domain information. To learn sentiment-wise discriminative representations in the absence of the target labels, we propose to learn domain invariance via contrastive learning (CL). In general, CL benefits from the definition of the augmented positive and negative pairs by treating instances as classes \citep{chen2020simple,khosla2020supervised,he2020momentum,chen2020improved}. Furthermore, the contrastive loss encourages the positive pairs to be close to each other and negative pairs to be far apart. Specifically, maximizing the similarities between positive pairs learns an invariant instance-based representation, and minimizing the similarities between negative pairs learns a uniformly distributed representation from a global view, making instances gathered near the task decision boundary away from each other \citep{Arora2019ATA,Grill2020BootstrapYO}. This will help to enhance task discrimination of the learned model.

For positive pairs, intuitively, we hope that the model could encode the original sentence and most domain-challenging examples to be closer in the representation space, gradually pulling examples to the domain decision boundary as training progresses.
For negative sampling, it widens the sentiment decision boundary and promotes better sentiment-wise discriminative features for both domains.
However, for cross-domain negative sampling, the contrastive loss may push the negative samples in the target (source) domain away from the anchor in the source (target) domain 
(see Fig.~\ref{fig:sampling} (b) left). This is contradictory to the objective of domain puzzles which try to pull different domains closer. To avoid the detriment of cross-domain repulsion, excluding samples with different domains from the negative set is of great importance. Therefore, we write the following contrastive infoNCE loss \cite{chen2020simple} as follow:
\begin{equation}
    \!\!\small
    \Loss_{{\rm contrast}}=-\frac{1}{N}\sum_{i}^{N}\log\frac{\exp(s(z_{i},z^{\prime}_{i})/\tau)}{\sum\nolimits_{k}^{N} \mathbbm{1}_{k\neq i}\exp(s(z_{i},z_{k})/\tau)},\!
\end{equation}
where $N$ is the mini batch size with samples from the same domain, $z_{i}=g(f(x_{i};\theta_{f}))$, and $g(\cdot)$ is one hidden layer projection head. We denote $x^{\prime}=x+\delta$ as the domain puzzle augmentation, $s(\cdot)$ computes cosine similarity, $\mathbbm{1}_{k\neq i}$ is the indicator function, and $\tau$ is the temperature hyperparameter.

\subsection{Consistency Regularization}
Given perturbed embedding $x+\delta$, which is crafted based on domain classification, we also encourage the model to produce consistent sentiment predictions with that of the original instance $f(x;\theta_{f},\theta_{y})$. 
For this, we minimize the symmetric KL divergence, which is formulated as:
\begin{equation}
    \Loss_{{\rm consist}}=\Loss(f(x;\theta_{f},\theta_{y}),f(x+\delta;\theta_{f},\theta_{y})).
\end{equation}

For overall training objective, we train the neural network in an end-to-end manner with a weighted sum of losses as follows. 
\begin{equation}
\small
\begin{split}
    \!\!\!&\min_{\theta_{f},\theta_{y},\theta_{d}}\sum_{(x,y)\sim \Dataset^{S}}\Loss(f(x;\theta_{f},\theta_{y}),y)+\\
    \!\!\!&\sum_{(x,y)\sim \Dataset^{S},\Dataset^{T}}[\alpha\Loss_{{\rm domain}}+\lambda \Loss_{{\rm contrast}}+\beta\Loss_{{\rm consist}}].
\end{split}
\label{loss}
\end{equation}
Details of proposed \method are summarized in Algorithm~\ref{algo:main}.

\begin{algorithm}[tb!]
\caption{{\method}}\label{algo:main}
\begin{algorithmic}[1]
	\INPUT For simplicity, $\theta$ is the parameter of the whole model. $T$: the total number of iterations, $(x, y)\sim\Dataset^{S}$: source dataset with sentiment label $y$, $(x,d)\sim\Dataset^{S}\Dataset^{D}$: source and target dataset with domain label $d$, $K$: the number of iterations for updating $\delta$, $\sigma^2$: the initialized variance, $\epsilon$: perturbation bound, $\eta$: the step size, $\gamma$: global learning rate, $N$: batch size, $\tau$: temperature, $g(\cdot)$:one hidden layer projection head. $\alpha_{adv}$, $\alpha$, $\lambda$ and $\beta$: weighting factor.  

	\For{{\small ${\rm epoch}=1,..,T$}}
	\For{minibatch $N$}
	    \State {\small$\delta\leftarrow\cN(0,\sigma^2I)$}
    	\For{{\small $m=1,..,K$}}
    	    \State {\small$g^{adv}_{d} \leftarrow \nabla_{\delta} \Loss(f(x + \delta; \theta), d)$}
    	    \State {\small $\delta\leftarrow\Pi_{\lVert \delta \rVert _{F}\leq \epsilon} (\delta+\eta g^{adv}_{d}/\lVert g^{adv}_{d} \rVert_{F})$}
    	\EndFor
    	\State {\small $\Loss_{{\rm domain}} \leftarrow \Loss(f(x;\theta),d)$}
    	
    	{\small$\qquad\qquad\quad+\alpha_{adv}\Loss(f(x+\delta;\theta),d)$}
    	
    	\State {\small $z=g(f(x;\theta))$}
    	\State {\small $z^{\prime}=g(f(x+\delta;\theta))$}
    	\For{{\small$i=1,...,N$ and $j=1,...,N$}}
    	    \State {\small$s^{\prime}_{i}=z_{i}^{\top}z^{\prime}_{j}/\lVert z_{i}\rVert\lVert z^{\prime}_{j}\rVert$}
    	    \State {\small $s_{i,j}=z_{i}^{\top}z_{j}/\lVert z_{i}\rVert\lVert z_{j}\rVert$}
    	\EndFor
    	\State {\small$\Loss_{{\rm contrast}} \leftarrow -\frac{1}{N}\sum\limits_{i}^{N}\log\frac{\exp(s^{\prime}_{i}/\tau)}{\sum_{j}^{N} \mathbbm{1}_{j\neq i}\exp(s_{i,j}/\tau)}$}
    	
    	\State {\small$\Loss_{{\rm consist}} \leftarrow \Loss(f(x;\theta),f(x+\delta;\theta))$}
    	
	    \State {\small $g_{\theta} \leftarrow \nabla_{\theta} \Loss(f(x; \theta), y)+\alpha \nabla_{\theta}\Loss_{{\rm domain}}$}
	    
	    {\small $\qquad\quad+ \lambda\nabla_{\theta}\Loss_{{\rm contrast}}+\beta\nabla_{\theta}\Loss_{{\rm consist}}$}

	    \State {\small $\theta \leftarrow \theta - \gamma g_{\theta}$}
	\EndFor
	\EndFor
		\OUTPUT $\theta$
\end{algorithmic}
\end{algorithm}

%% file: 004experiments.tex
 \subsection{Datasets}
 \textbf{Amazon Benchmark} \citep{blitzer2007biographies}\footnote{https://www.cs.jhu.edu/~mdredze/datasets/sentiment/}. We conduct experiments on this dataset for completeness since most of previous works report results on it. The dataset contains four domains: Book (BK), DVD (D), Electronics (E) and Kitchen housewares (K). There are 2,000 balanced labeled data for each domain, we randomly select 20,000 unlabeled reviews for BK, D and E. For K, only 17,856 unlabeled reviews are available.
 
 \noindent\textbf{Amazon review dataset} \citep{he2018adaptive}\footnote{https://github.com/ruidan/DAS}. This dataset considers neutral instances which may not bias the dataset and bring more challenges\footnote{The original crawled reviews contain star ratings (1 to 5 stars). Comparing with Amazon Benchmark which discards the neutral class, this dataset labels them with rating $<3$, $>3$, $=3$ as negative, positive, and neutral respectively.}. The dataset also contains four domains: Book (BK), Electronics (E), Beauty (BT), and Music (M). Following \citet{he2018adaptive}, we treat set $1$ as labeled dataset containing $6,000$ instances, and treat set $2$ as unlabeled dataset which also contains $6,000$ instances. 
More details about two datasets can be found in Appendix \ref{sec:appendixA}. 
 
\subsection{Experiment Settings}
 For unsupervised adaptation setting, we should not have access to target labeled data at the training phase, so trained models with minimum classification error on the source validation set is saved for evaluation.  At this point, we suppose a good model that generalizes well on the target domain is able to reach high performance on both validation and test set at the same time. We evaluate our model with $5$ runs in all experiments, and we report the average score, standard deviation and paired t-test results. 
 
 \subsection{Implementation Details}
 For pre-trained language model, we use BERT base uncased \citep{Devlin2019BERTPO} as the basis for all experiments. The max length is set to $512$. For optimizer, we use AdamW \citep{Kingma2015AdamAM,loshchilov2018decoupled} with weight decay 0.01 (for BERT baseline, we set 1e-4). We set the learning rate as 1e-5, and we use a linear scheduler with warm-up steps $0.1$ of total training steps.
 
 We set the number of adversarial iterations to be 1, adversarial weighting factor $\alpha_{adv}$ = 1 and we use $l_{2}$ norm to compute projections. We also follow \citet{zhu2020freelb} to set other adversarial hyperparameters, e.g., adversarial step size $\eta$ = 5e-2, perturbation bound $\epsilon$ = 5e-2. For weighting factors, we set $\alpha$ = 1e-3, $\lambda$ = 3e-2 and $\beta$ = 5. We train the model with $8$ epochs, temperature $\tau$ and batch size $N$ will be discussed later. Each adaptation requires half one hour on one A-100.
  \begin{table*}[t!]
\centering
\resizebox{\textwidth}{!}
{
\setlength{\tabcolsep}{0.8mm}
{
\begin{tabular}{l|llllllllllll|l}
\toprule
\hline
\textbf{Method} & E$\to$BK            & BT$\to$BK & M$\to$BK & BK$\to$E & BT$\to$E & M$\to$E & BK$\to$BT & E$\to$BT & M$\to$BT & BK$\to$M & E$\to$M & BT$\to$M & Ave. \\ 
\hline 
BERT base & 64.34$_{1.3}^{*}$ & 65.87$_{1.9}^{*}$ & 64.12$_{1.7}^{*}$ & 52.25$_{1.8}^{*}$ & 66.01$_{1.0}^{*}$ & 59.49$_{1.1}^{*}$ & 55.01$_{1.7}^{*}$ & 66.02$_{1.3}^{*}$ & 56.50$_{2.1}^{*}$ & 55.66$_{1.5}^{*}$ & 61.34$_{0.8}^{*}$ & 62.34$_{1.2}^{*}$ & 60.74$_{1.5}$ \\ 
\cdashline{1-14}
KL  & 64.77$_{0.8}^{*}$ & 67.03$_{0.7}^{*}$ & 65.13$_{1.0}^{*}$ & 56.96$_{1.2}^{*}$ & 65.43$_{1.5}^{*}$ & 60.30$_{0.7}^{*}$ & 58.47$_{0.9}^{*}$ & 66.22$_{0.5}^{*}$ & 58.80$_{1.3}^{*}$ & 57.53$_{0.7}^{*}$ & 60.02$_{1.4}^{*}$ & 64.41$_{1.1}^{*}$ & 62.09$_{1.0}$\\

MMD  & 65.41$_{0.7}^{*}$ & 68.54$_{1.1}^{*}$ & 64.77$_{1.3}^{*}$ & 58.84$_{1.6}^{*}$ & 66.39$_{1.5}^{*}$ & 59.29$_{1.9}^{*}$ & 60.75$_{1.5}^{*}$ & 66.50$_{0.9}^{*}$ & 57.42$_{1.3}^{*}$ & 59.37$_{1.4}^{*}$ & 59.47$_{1.0}^{*}$ & 65.03$_{1.6}^{*}$ & 62.65$_{1.3}$ \\

DANN  & 66.43$_{2.4}^{*}$ & 67.74$_{2.9}^{*}$ & 65.64$_{3.7}^{*}$ & 54.31$_{4.1}^{*}$ & 65.93$_{3.0}^{*}$ & 59.91$_{1.9}^{*}$ & 58.11$_{3.5}^{*}$ & 67.19$_{0.5}^{*}$ & 56.00$_{2.7}^{*}$ & 54.02$_{2.9}^{*}$ & 60.36$_{3.1}^{*}$ & 63.09$_{2.7}^{*}$ & 61.56$_{2.8}$\\

back-trans+CL  & 67.16$_{1.4}^{*}$ & 68.31$_{0.8}^{*}$ & 67.78$_{1.8}^{*}$ & 59.55$_{1.1}^{*}$ & 66.51$_{0.3}^{*}$ & 60.13$_{1.4}^{*}$ & 60.92$_{0.7}^{*}$ & 68.54$_{0.8}^{*}$ & 60.44$_{0.9}^{*}$ & 60.21$_{1.1}^{*}$ & 61.73$_{1.3}^{*}$ & 64.29$_{0.7}^{*}$ & 63.79$_{1.0}$ \\

mask  & 65.17$_{1.1}^{*}$ & 66.91$_{1.4}^{*}$ & 65.59$_{0.9}^{*}$ & 52.74$_{1.5}^{*}$ & 67.14$_{1.9}^{*}$ & 60.48$_{0.7}^{*}$ & 54.76$_{1.6}^{*}$ & 66.50$_{1.4}^{*}$ & 56.61$_{1.7}^{*}$ & 55.64$_{1.6}^{*}$ & 62.17$_{0.9}^{*}$ & 63.09$_{1.0}^{*}$ & 61.40$_{1.3}$ \\

mask+CL  & 68.80$_{1.2}^{*}$ & 69.93$_{0.7}^{*}$ & 70.65$_{0.9}^{*}$ & 60.50$_{1.3}^{*}$ & 68.02$_{0.7}$ & 60.80$_{1.2}^{*}$ & 62.17$_{0.8}^{*}$ & 69.17$_{0.5}$ & 62.06$_{1.5}^{*}$ & 60.56$_{1.4}^{*}$ & 63.94$_{0.7}$ & 65.40$_{0.6}$ & 65.17$_{0.9}$ \\

\hline
\method  & \textbf{70.33}$_{0.3}$ & \textbf{70.92}$_{0.6}$ & \textbf{71.11}$_{0.7}$ & \textbf{62.36}$_{0.7}$ & \textbf{68.41}$_{0.2}$ & \textbf{62.11}$_{0.6}$ & \textbf{64.13}$_{1.4}$ & \textbf{69.33}$_{0.4}$ & \textbf{65.40}$_{0.8}$ & \textbf{64.67}$_{1.7}$ & \textbf{64.67}$_{1.0}$ & \textbf{66.70}$_{1.4}$ & \textbf{66.68}$_{0.8}$\\

\bottomrule

\end{tabular}
}
}
\caption{Accuracy (\%) results of Amazon review dataset. For example, E$\to$BK denotes training on Electronics (E) and adapting to Book (BK). All the results are reported with average and standard deviation in 5 runs. $*$ indicates the DCCL improvements are significant with $p<0.05$. There are four domains available in the dataset, therefore we have 12 adaptation groups of tasks.}
\label{table:results}
\end{table*}
 \begin{table*}[t]
\centering
\resizebox{\textwidth}{!}
{
\setlength{\tabcolsep}{0.8mm}
{
\begin{tabular}{l|llllllllllll|l}
\toprule
\hline
\textbf{Method} & D$\to$BK & K$\to$BK & E$\to$BK & BK$\to$D & K$\to$D & E$\to$D & BK$\to$K & D$\to$K & E$\to$K & BK$\to$E & D$\to$E & K$\to$E & Ave. \\ 
\hline 
R-PERL & 85.6 & 83.0 & 83.9 & 87.8 & 85.6 & 84.8 & 90.2 & 90.4 & 91.2 & 87.2 & 89.3 & 91.2 & 87.5 \\ 

DAAT & 90.86 & 87.98 & 88.91 & 89.70 & 88.81 & 90.13 & 90.75 & 90.50 & \textbf{93.18} & 89.57 & 89.30 & 91.72 & 90.12 \\ 
\cdashline{1-14}

BERT base & 89.76$_{0.3}^{*}$ & 88.31$_{0.4}^{*}$ & 88.09$_{0.9}^{*}$ & 89.27$_{0.5}^{*}$ & 87.68$_{0.8}^{*}$ & 88.41$_{1.2}^{*}$ & 88.27$_{0.4}^{*}$ & 87.67$_{0.5}^{*}$ & 92.16$_{0.7}$ & 87.18$_{0.4}^{*}$ & 86.91$_{0.7}^{*}$ & 91.22$_{0.8}$ & 88.74$_{0.6}$\\ 

KL & 89.23$_{0.5}^{*}$ & 87.61$_{0.5}^{*}$ & 88.37$_{0.7}^{*}$ & 89.49$_{0.6}^{*}$ & 88.03$_{0.5}^{*}$ & 88.56$_{0.7}^{*}$ & 88.77$_{0.5}^{*}$ & 87.89$_{0.7}^{*}$ & 91.21$_{1.3}$ & 88.62$_{0.4}^{*}$ & 87.03$_{0.9}^{*}$ & 90.34$_{1.5}^{*}$ & 88.76$_{0.7}$ \\ 

MMD & 88.68$_{0.7}^{*}$ & \textbf{88.61}$_{0.3}$ & 88.27$_{1.0}^{*}$ & 88.38$_{1.0}^{*}$ & 89.09$_{0.5}$ & 89.31$_{0.6}^{*}$ & 87.74$_{0.8}^{*}$ & 89.04$_{0.8}^{*}$ & 91.20$_{1.0}$ & 89.49$_{0.3}^{*}$ & 88.70$_{0.5}^{*}$ & 89.79$_{1.2}^{*}$ & 89.03$_{0.6}$ \\ 

DANN & 88.64$_{1.3}^{*}$ & 86.09$_{1.7}^{*}$ & 87.91$_{1.5}^{*}$ & 89.56$_{1.0}$ & 89.11$_{0.9}$ & 89.19$_{1.1}^{*}$ & 89.68$_{1.3}^{*}$ & 88.73$_{1.8}^{*}$ & 91.41$_{1.3}^{*}$ & 87.30$_{1.6}^{*}$ & 88.53$_{0.6}^{*}$ & 90.81$_{1.1}^{*}$ & 88.91$_{1.3}$ \\ 

mask+CL & 90.74$_{0.3}$ & 88.28$_{0.7}$ & 89.44$_{0.4}^{*}$ & \textbf{90.31}$_{0.2}$ & 88.74$_{0.5}^{*}$ & 89.72$_{0.8}^{*}$ & 90.40$_{0.4}^{*}$ & 89.57$_{0.4}^{*}$ & 92.49$_{0.4}$ & 89.91$_{0.3}^{*}$ & \textbf{89.65}$_{0.4}$& 91.19$_{0.6}$ & 90.08$_{0.5}$\\ 

\hline
\method & \textbf{91.17}$_{0.3}$ & 88.53$_{0.4}$ & \textbf{89.70}$_{0.5}$ & 90.03$_{0.4}$ & \textbf{89.53}$_{0.3}$ & \textbf{90.49}$_{0.5}$ & \textbf{91.05}$_{0.5}$ & \textbf{90.78}$_{0.4}$ & 92.54$_{0.5}$ & \textbf{90.42}$_{0.3}$ & 89.55$_{0.4}$ & \textbf{91.93}$_{0.5}$ & \textbf{90.48}$_{0.4}$ \\ 

\bottomrule

\end{tabular}
}
}
\caption{Accuracy (\%) results of Amazon Benchmark. Results of R-PERL and DAAT are taken from \citet{ben2020perl} and \citet{du2020adversarial} respectively.}
\label{table:bench_results}
\end{table*}

 \subsection{Baselines}
 \label{sec:baselines}
 \textbf{BERT base}: Fine-tune BERT on the source, without using target unlabeled data, then directly evaluate on the target labeled data. 
 \textbf{KL}: Use symmetric KL-divergence loss of embedded instances between the source and target domains \citep{zhuang2015supervised}. 
 \textbf{MMD}: Maximum Mean Discrepancy loss \citep{gretton2012kernel} measures the distance based on the notion of embedding probabilities in a reproducing kernel Hilbert space. We implement a gaussian kernel which is a common choice. 
 \textbf{DANN}: The adaptation rate is $\lambda=\frac{2}{1+\exp(-\gamma p)}-1$, $p=\frac{t}{T}$, where $t$ and $T$ are the number of current training steps and total steps. $\gamma$ requires careful tuning within $[0.05,0.1,0.15,0.2]$. 
 \textbf{back-trans+CL}: To investigate the effectiveness of domain puzzles, we implement Back translation \citep{sennrich2016improving,edunov2018understanding}, which is one of the widely used data augmentation techniques. We utilize en-de translation model pre-trained on WMT19 and released in fairseq \citep{ott2019fairseq}. The model is trained with contrastive loss on the source and target domain respectively. 
 \textbf{mask+CL}: We mask domain specific tokens to make the augmentation become domain-agnostic. Since information extraction is not our focus, we identify domain specific tokens via a simple frequency-ratio method \citep{li2018delete,Wu2019MaskAI}: $s(u,d)=\frac{{\rm count}(u,\Dataset^{d})+\lambda}{\sum_{d^{\prime}\in\Dataset,d^{\prime}\neq d}{\rm count}(u,\Dataset^{d^{\prime}})+\lambda}$, where ${\rm count}(u,\Dataset^{d})$ represents the number of times a token $u$ appears in domain $d$. Smoothing $\lambda$ is set to 1. When $s(u,d)$ is larger than $5$, we mark a token $u$ as a domain specific token. Through counting, the number of masked tokens accounted for 0.06 of the total length. 
 \textbf{mask}: To investigate the effectiveness of contrastive loss, after masking domain-specific tokens, we further let the model train with augmented data without contrastive loss.
 \textbf{R-PERL} \citep{ben2020perl} is a pivot-based method, and \textbf{DAAT} \cite{du2020adversarial} combines DANN and post-training. All the methods in Tabel \ref{table:results} and Table \ref{table:bench_results} are implemented based on BERT model for fair comparisons.
 
 \subsection{Results}
 \textbf{Will contrastive learning necessarily help the Unsupervised Domain Adaptation?} \\
 As discussed earlier, 
when performing contrastive learning on source labeled examples and target unlabeled examples respectively, it learns a uniformly distribute representation and helps promote better discriminative features for both domains. 
However, for some adaptation tasks in Table~\ref{table:results}, for example, BT$\to$E, M$\to$E, and E$\to$M, back-trans+CL shows that contrastive learning only gains marginal benefit. When masking domain-specific tokens and pulling original sentence representations to those domain puzzles, the effect of contrastive learning becomes more apparent (mask+CL with average score 65.17 compared with back-trans+CL 63.79 and mask 61.40 in Table.\ref{table:results}). This finding helps to explain that choices of positive examples are critical and domain confused augmentations will further benefit adaptation.

 \vspace{1mm}
 \noindent\textbf{\method outperforms baselines.} \\
 From Table \ref{table:results} we can observe that the proposed \method outperforms all other methods with a large margin and $p<0.05$ using paired t-test, and 5.94\% improvement over BERT base. From Table \ref{table:bench_results}, we can also observe 1.74\% improvement over BERT base, and \method also surpasses state-of-the-art methods R-PERL \citep{ben2020perl} and DAAT \citep{du2020adversarial}. We note that Amazon Benchmark dataset is quite easy, since it discards neutral instances, BERT base model has already achieved high scores on this dataset. Besides, we observe that the effect of distribution matching methods (KL and MMD) is limited on two datasets. The reason might be that pre-trained language models trained with massive and heterogeneous corpora already have strong generalization ability. Learning such cross-domain and instance-based matching will bring perplexity to language models and sacrifice task discrimination \citep{saito2017asymmetric,lee2019drop}. On the contrary, the proposed \method retains such information in a self-supervised way. Furthermore, we notice that DANN is very unstable, besides adaptation rate $\lambda$, the model is also sensitive to other hyperparameters such as learning rate and training epochs because the performance on the target domain will keep decreasing when training with longer steps. Hence, it is difficult for the model to achieve the lowest error rates on both the source and target domains simultaneously. Compared to DANN, \method is much more stable and has lower standard deviations on most adaptation tasks.
 
 \begin{figure}[ht]
     \centering
     \includegraphics[width=0.9\columnwidth]{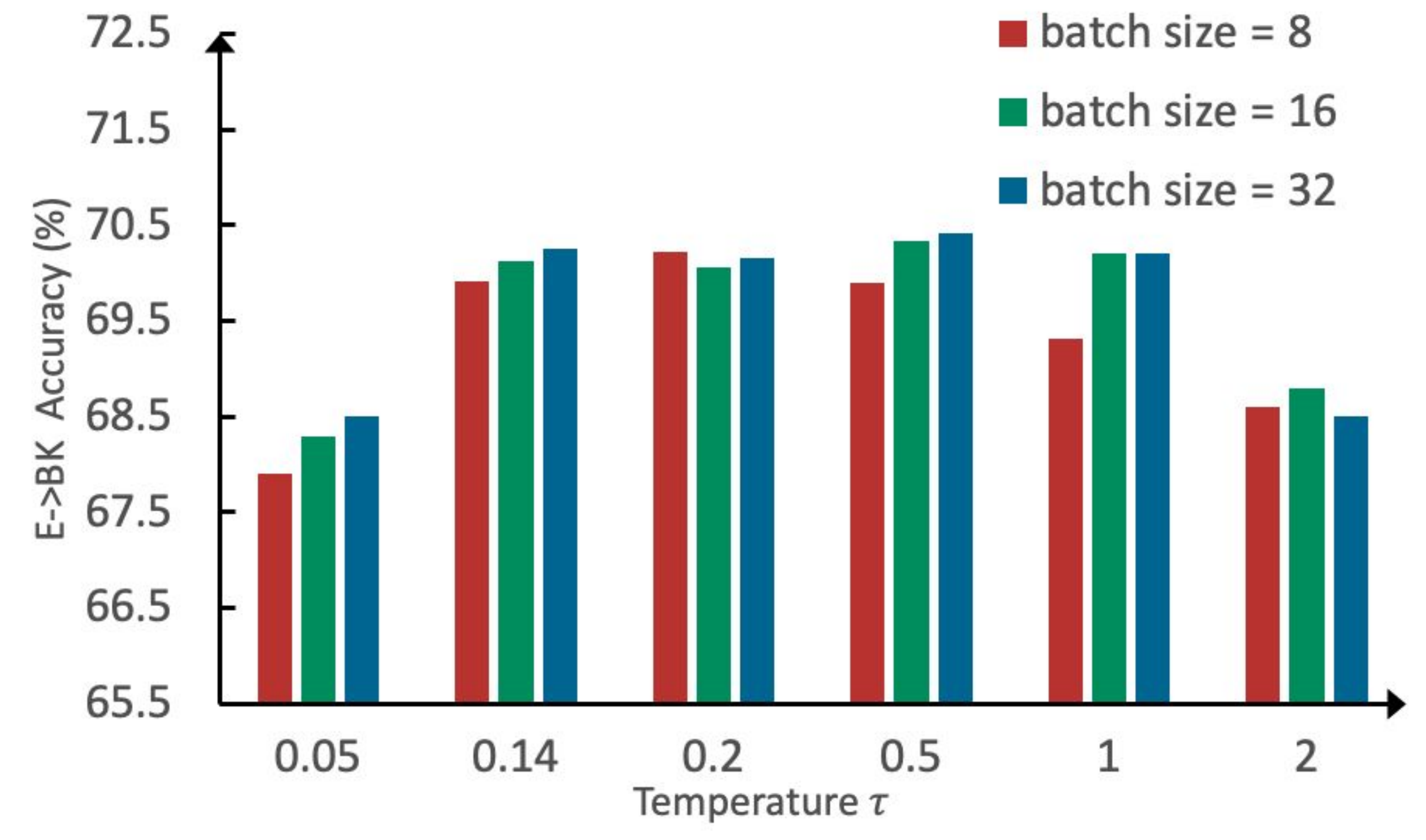}
     \caption{Hyperparameters for contrastive learning}
     \label{fig:tem}
 \end{figure}
 
\vspace{1mm}
\noindent\textbf{Contrastive learning designs.} \\
We explore different hyperparameter values for the proposed \method in E$\to$BK, as is shown in Fig.~\ref{fig:tem}. We find that a temperature of 0.5 combined with the batch size 32 can achieve the best performance. We also notice that setting the temperature too high or too low can significantly affect adaptation performances, while a larger batch size usually brings a relatively smaller improvement.
 
\subsection{Ablation Studies}
 We conduct an ablation study on each component in Eq. \ref{loss} to inspect the contribution of contrastive learning, as is shown in Table \ref{table:ablation}. We can see that every single component can render a better adaptation performance. In particular, the effectiveness of $\Loss_{{\rm contrast}}$ is observable with 3\textasciitilde 5 performance gain compared to baselines (row 4 vs. row 2, and row 5 vs. row 3). When training curated domain puzzles as simple augmentations without contrastive loss, we can observe only a slight improvement (row 2 vs. row 1). This result demonstrates that the performance gain brought by \method does not come from data augmentation, and learning robustness against adversarial attacks will not largely help adaptation.
 
 \begin{table}[h]
\centering
\resizebox{\columnwidth}{!}{
\begin{tabular}{lll|ll}
\toprule
 $\Loss_{{\rm domain}}$ & $\Loss_{{\rm consist}}$ & $\Loss_{{\rm contrast}}$ & E$\to$BK & M$\to$BT \\ \hline
  &  &  & 64.65 & 55.85 \\
 $\surd$ &  &  & 67.23 & 58.9 \\
 $\surd$ & $\surd$ &  & 67.85 & 59.1 \\
 $\surd$ &  & $\surd$ & 70.12 & 64.73 \\
 $\surd$ & $\surd$ & $\surd$ & \textbf{70.21} & \textbf{64.87} \\
\bottomrule
\end{tabular}
}
\caption{Ablation studies of \method on each component}
\label{table:ablation}
\end{table}

%% file: 005analysis.tex
\subsection{Visualization}
\begin{figure}[htbp]
    \centering
    \includegraphics[width=0.9\columnwidth]{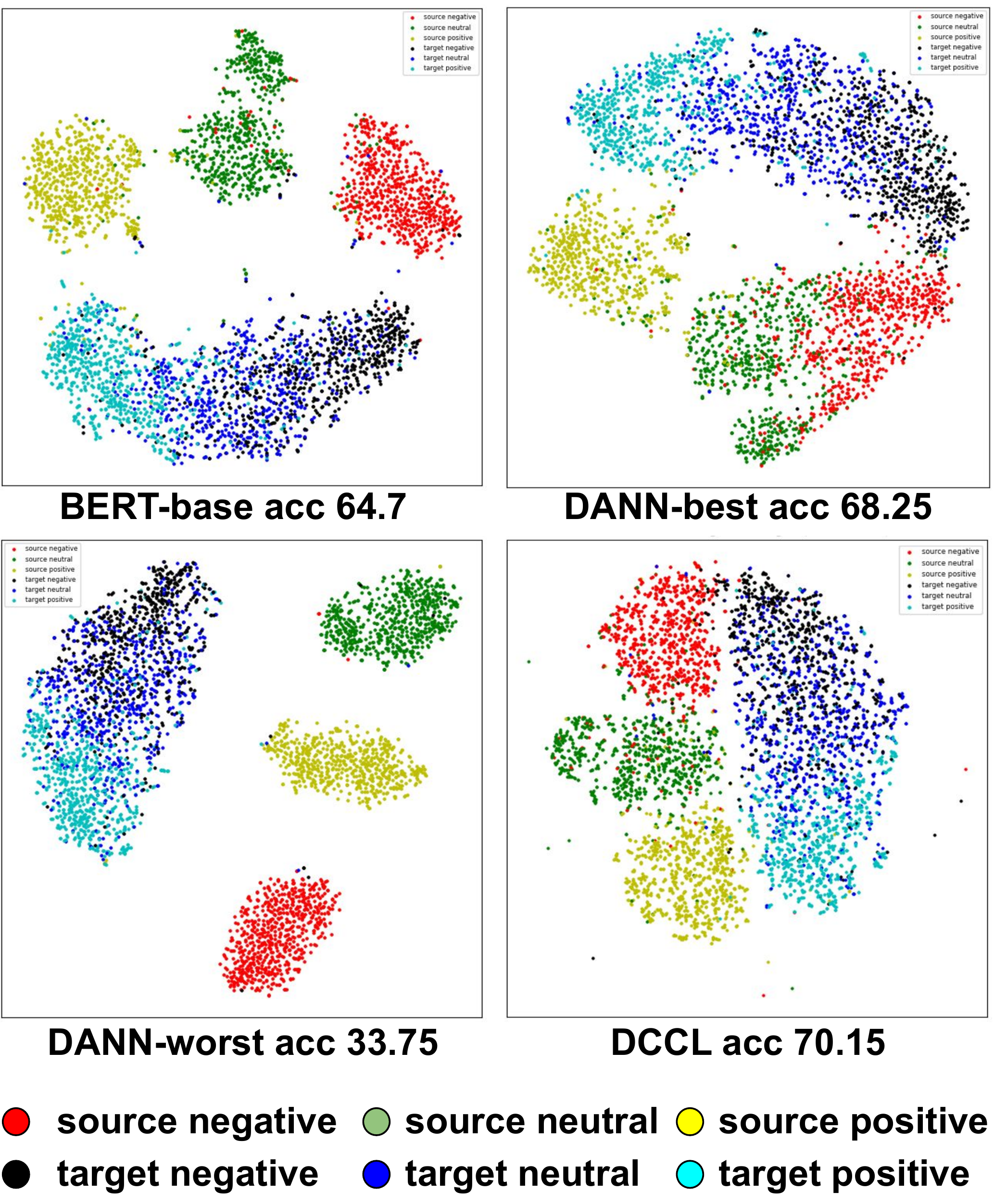}
    \caption{t-SNE visualization for E$\to$BK task}
    \label{fig:tsne}
\end{figure}

We perform visualizations for trained representations as illustrated in Fig.~\ref{fig:tsne}. When training with the source domain and then adapting to the target domain (BERT-base), we can observe a considerable domain shift for BERT encoder on this amazon review dataset. Moreover, as mentioned before, continuing training DANN with larger epochs will substantially drop the score (from the highest point (DANN-best) to the lowest point (DANN-worst)). However, we can also see that \method mitigates domain shift but remains good sentiment discrimination on the target domain.

\subsection{Quantitative Results}
\begin{figure}[htbp]
    \centering
    \includegraphics[width=0.85\columnwidth]{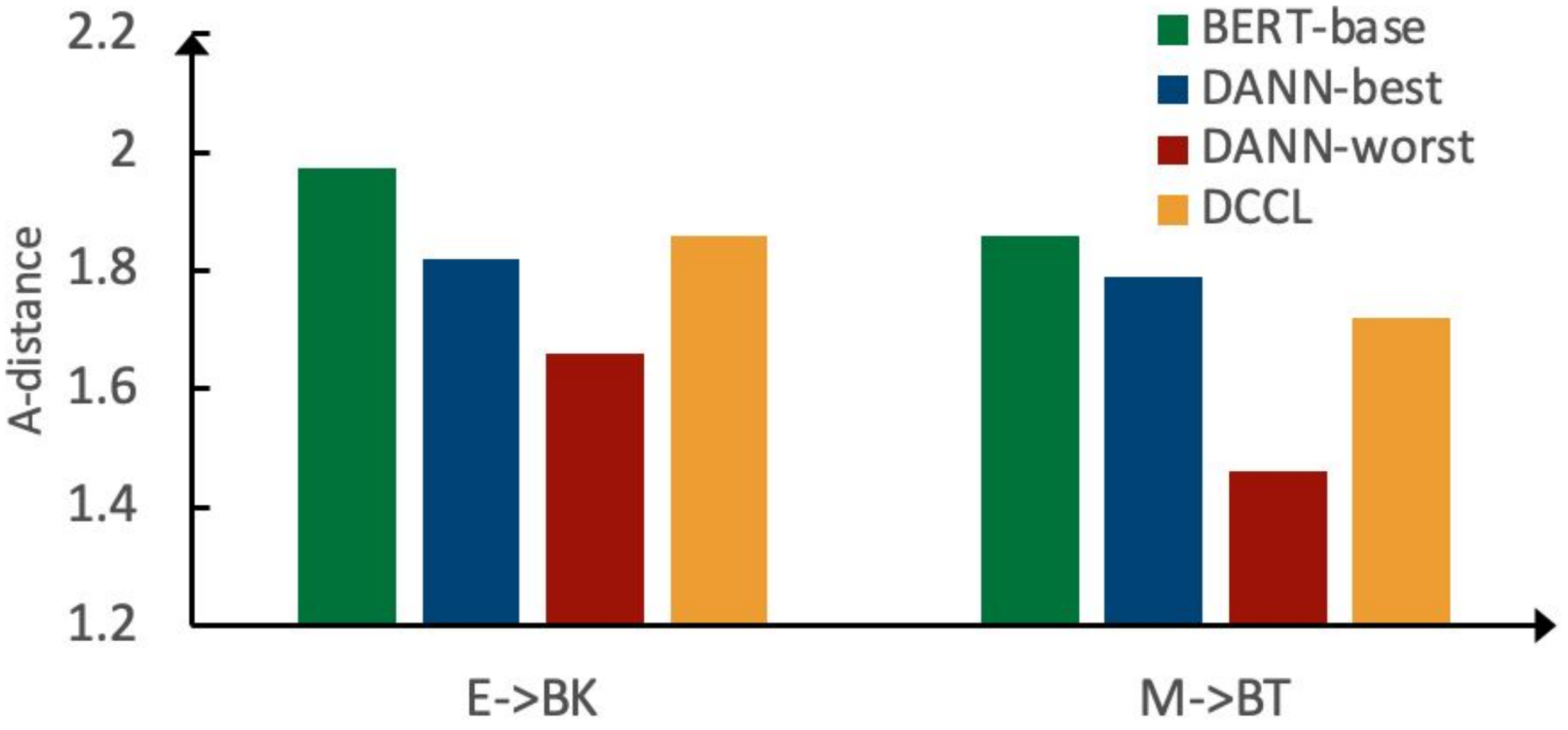}
    \caption{$\mathcal{A}$-distance for E$\to$BK and M$\to$BT tasks}
    \label{fig:a-dis}
\end{figure}

$\mathcal{A}$-distance measures domain discrepancies \citep{ben2007analysis,ben2010theory}, with the definition as $d_{\mathcal{A}}=2(1-2\epsilon)$, where $\epsilon$ is the domain classification error. To fairly compare with $\mathcal{A}$-distance of the baselines, we use linear SVM to calculate $\epsilon$ following previous work \citep{saito2017asymmetric,du2020adversarial}. We randomly select 2,000 instances for both source and target domain and split them with 1:1 for train and test for the SVM model. From Fig.~\ref{fig:a-dis}, we can observe that \method can learn a good balance between sentiment classification and domain discrepancy, compared to DANN-best and DANN-worst.

%% file: 006related.tex
\textbf{Unsupervised Domain Adaptation} \\
UDA in NLP has the following approaches: (1) Pivot-based methods use unlabeled data from both domains, trying to discover characteristics that are similar \citep{pan2010cross}. Some recent works extend pivots with autoencoders and contextualized embeddings \citep{ziser2017neural,miller2019simplified,ben2020perl}. 
(2) Pseudo-labeling leverages a trained classifier to predict labels on unlabeled examples, which are subsequently considered as gold labels \citep{yarowsky1995unsupervised,zhou2005tri,mcclosky2006reranking}. Recent works also combine this technique with pre-trained language models \citep{lim2020semi,ye2020feature}.
(3) Data selection methods adopt domain similarity metrics to find the best match for each data and use curriculum learning for large pre-trained models \citep{ma2019domain,aharoni2020unsupervised}. A recent method adopt distance-bandit \citep{guo2020multi} for similarity metric. 
(4) Domain Adversarial Neural Networks \citep{ganin2016domain}. Some approaches leverage Wasserstein distance to stabilize adversarial training \citep{shen2018wasserstein,shah2018adversarial}, and combine it with post-training which can produce better adversarial results \citep{du2020adversarial}. 
(5) Adaptive pre-training is a more straightforward but effective method \citep{gururangan2020don,han2019unsupervised,karouzos2021udalm} by leveraging the objective of masked language model (MLM). 
A wide range of pre-training methods for domain adaptation (multi-phase, multi-task) are put forward \citep{han2019unsupervised,karouzos2021udalm}. 

\vspace{2mm}
\noindent\textbf{Contrastive learning} \\
CL has recently gained popularity as a reliable approach for unsupervised representation learning. 
It is generally acknowledged that a good representation should distinguish itself from other instances while identifying similar instances. 
For Computer Vision, there are approaches obtaining augmented images using transformations including cropping, rotation, etc. \citep{chen2020simple,khosla2020supervised,he2020momentum,chen2020improved}. As for Natural Language Processing, many works study different label-preserving augmentations, such as back-translations, synonyms, adversaries, dropout, and their combinations \citep{qu2021coda,gao2021simcse}. In addition, many pre-trained language models trained with contrastive loss are also released. DeCLUTR \citep{giorgi2021declutr} and CLEAR \citep{wu2020clear} jointly train the model with a contrastive objective and a masked language model setting. ConSERT \citep{Yan2021ConSERTAC} overcomes the collapse problem of BERT-derived sentence representations and makes them more suitable for downstream applications by using unlabeled texts.

%% file: 007conclusion.tex
In this work, we put forward a new concept, domain puzzles, which can be crafted through domain-specific token mask and domain-confused adversarial attacks. And we provide a more stable and effective solution to learn domain invariance for unsupervised domain adaptation. The proposed method \method surpasses baselines with a large margin by mitigating domain shift without losing discriminative power on the target domain. Moreover, the proposed framework can also be extended to other NLP tasks demanding adaptations, and we leave this for future work.

%% file: 008acknowledgement.tex
This work is supported by the 2020 Microsoft Research Asia collaborative research grant.

%% file: 010appendixA.tex
We obtain the Amazon review datasets from \citet{he2018adaptive}. This dataset does not remove neutral labels and will not be problematic in UDA situation where the label information of the target domain is not available. In addition, reserving neutral labels also bring challenges for pre-trained language model, making it more favorable for self-supervised representation learning. Summary of this dataset is availble in Table \ref{table:data_}.

\begin{table}[h]
\centering
\resizebox{\columnwidth}{!}{
\begin{tabular}{l|llll|l}
\toprule
Domain                       & \multicolumn{1}{c}{} & \#Neg & \multicolumn{1}{c}{\#Neu} & \#Pos & Total \\ \hline
\multirow{2}{*}{Book}        & Set 1                & 2000  & 2000                      & 2000  & 6000  \\
                             & Set 2                & 513   & 663                       & 4824  & 6000  \\ \hline
\multirow{2}{*}{Electronics} & Set 1                & 2000  & 2000                      & 2000  & 6000  \\
                             & Set 2                & 694   & 489                       & 4817  & 6000  \\ \hline
\multirow{2}{*}{Beauty}      & Set 1                & 2000  & 2000                      & 2000  & 6000  \\
                             & Set 2                & 616   & 675                       & 4709  & 6000  \\ \hline
\multirow{2}{*}{Music}       & Set 1                & 2000  & 2000                      & 2000  & 6000  \\
                             & Set 2                & 785   & 774                       & 4441  & 6000  \\
\bottomrule
\end{tabular}
}
\caption{Amazon review dataset}
\label{table:data_}
\end{table}

Each domain contains two sets, set 1 contains 6000 instances with balanced class labels, and set 2 contains instances that are randomly sampled from the larger dataset \citep{mcauley2015image}, preserving authentic label distribution, examples in these two datasets do not overlap. Following \cite{he2018adaptive}, we use set 1 from the source domain as the training set for all our experiments. Since label distribution in the target domain is unpredictable and out of control in real life, so it's more reasonable to use set 2 from the target domain as the unlabeled set, lastly the model will be evaluated in set 1 from target domain. For data split, we randomly sample $1000$ instances from the source labeled dataset as validation set. When running UDA experiments, the model will train on $5000$ source labeled examples and $6000$ target unlabeled examples, then validate on $1000$ source labeled examples.

For Amazon Benchmark\citep{blitzer2007biographies}, it also contains four domains: Book (BK), DVD (D), Electronics (E) and Kitchen housewares (K). There are 2000 balanced labeled data for each domain, we randomly select 20000 unlabeled reviews for BK, D and E. For K,  only 17856 unlabeled reviews are available, statistics of Amazon Benchmark can be find in Table.\ref{table:data__}. For data split, 1600 balanced samples are randomly sampled from the source labeled dataset, and 400 for validation.

\begin{table}[h]
\centering
\resizebox{\columnwidth}{!}{
\begin{tabular}{l|ll}
\toprule
Domains               & Labeled & Unlabeled \\ \hline
Book                  & 2000    & 973194    \\ \hline
Dvd                   & 2000    & 122438    \\ \hline
Electronics           & 2000    & 21009     \\ \hline
Kitchen \& housewares & 2000    & 17856   \\
\bottomrule
\end{tabular}
}
\caption{Amazon Benchmark statistics}
\label{table:data__}
\end{table}

%% file: 011appendixB.tex
Domain Adversarial Neural Network \citet{ganin2016domain} proposes Domain Adversarial Neural Network (DANN), which learns domain invariant and discriminative features simultaneously. This approach is motivated by the idea that an adaptation algorithm could learn good representations for cross-domain transfer if it cannot differentiate the domain of the input observations. The optimization objective is:
\begin{align}
    &\min_{\theta_{f},\theta_{y}}\sum_{(x,y)\sim \Dataset^{S}}[\Loss(f(x;\theta_{f},\theta_{y}),y) + \lambda R_{\theta_{f}}], \label{eq11}\\
    &R_{\theta_{f}}=\max_{\theta_{d}}\sum_{(x,d)\sim \Dataset^{S},\Dataset^{T}}[-\Loss(f(x;\theta_{f},\theta_{d}),d)] \label{eq12},
\end{align}
where $\theta_{d}$ is the parameter corresponding to the computation of the domain classification, $d$ is the domain label, $R_{\theta_{f}}$ is a regularizer weighted by $\lambda$. Objective (\ref{eq11}) learns task discrimination by minimizing task classification loss and tries to make features similar across domains by maximizing the domain classification loss. 

Although the domain classifier with parameters $\theta_{d}$ could perfectly classify different domains, the balance between two terms in (\ref{eq11}) is hard to maintain. Hence, the training process becomes unstable, requiring an elaborate adaptation rate $\lambda$ tuning \citep{shen2018wasserstein, shah2018adversarial,du2020adversarial}. Furthermore, the encoder could learn trivial solutions \citep{karouzos2021udalm} which produce features with flipped domain predictions.